\documentclass[11pt]{article}
\usepackage{foqa}
\usepackage{times}
\usepackage{url}
\usepackage{latexsym}

\usepackage{amsmath}
\usepackage[T1]{fontenc}
\usepackage{lmodern}
\usepackage{hyperref}
\usepackage{footnote}
\usepackage{booktabs}
\usepackage{tikz}

\newcommand{\githuburl}{\url{https://github.com/alexandrainst/foqa}}
\newcommand{\hfurl}{\url{https://huggingface.co/datasets/alexandrainst/foqa}}
\newcommand{\appurl}{\url{https://huggingface.co/spaces/saattrupdan/foqa-validation}}
\newcommand{\wikiurl}{\url{https://hf.co/datasets/alexandrainst/scandi-wiki}}

\aclfinalcopy

\title{FoQA: A Faroese Question-Answering Dataset}

\author{
    Annika Simonsen \\
    University of Iceland \\
    {\tt ans72@hi.is} \\\And
    Dan Saattrup Nielsen \\
    Alexandra Institute \\
    {\tt dan.nielsen@alexandra.dk} \\\And
    Hafsteinn Einarsson \\
    University of Iceland \\
    {\tt hafsteinne@hi.is}
}

\date{}

\begin{document}

\maketitle
\begin{abstract}
  We present FoQA, a Faroese extractive question-answering (QA) dataset with 2,000
  samples, created using a semi-automated approach combining Large Language Models
  (LLMs) and human validation. The dataset was generated from Faroese Wikipedia articles
  using GPT-4-turbo for initial QA generation, followed by question rephrasing to
  increase complexity and native speaker validation to ensure quality. We provide
  baseline performance metrics for FoQA across multiple models, including LLMs and BERT,
  demonstrating its effectiveness in evaluating Faroese QA performance. The dataset is
  released in three versions: a validated set of 2,000 samples, a complete set of all
  10,001 generated samples, and a set of 2,395 rejected samples for error analysis.
\end{abstract}

\section{Introduction}
\label{sec:introduction}

Recent NLP advancements, driven by the transformer
architecture~\cite{vaswani2017attention}, have led to large-scale models that excel in
understanding~\cite{devlin2018bert} and generating~\cite{brown2020language} human
language. While many models are ``massively
multilingual''~\cite{conneau2019unsupervised,he2021debertav3,brown2020language} they
often perform better on high-resource languages, leaving low-resource languages
under-supported. Furthermore, low-resource languages typically have limited access to
native speakers who can serve as data annotators, making it difficult to create
high-quality evaluation datasets. High-quality evaluation datasets are crucial for
assessing and improving models for these languages, helping to measure performance and
guide language technology development.

Extractive QA datasets~\cite{srivastava2024towards} are especially useful, as they
simulate real-world applications like retrieval-augmented
generation~\cite{gao2023retrieval}. Creating these datasets traditionally requires
substantial human effort, often involving multiple annotators for question generation
and answer validation. Standardising methods for creating these datasets can
significantly advance technology for low-resource languages.

Our research addresses these challenges and makes the following key contributions:

\begin{itemize}
    \item \textbf{An efficient, single-annotator methodology for producing high-quality
      extractive QA datasets} using a semi-automated approach that significantly reduces
      the human effort required for dataset creation, provided as an open-source Python
      codebase\footnote{\githuburl}
    \item \textbf{The first extractive QA dataset for Faroese} using this
      method\footnote{\hfurl}.
\end{itemize}

\section{Related Work}
\label{sec:related_work}

QA systems are divided into extractive and abstractive types~\cite{fan-etal-2019-eli5}.
This work focuses on extractive QA, also known as reading comprehension, where text
passages are paired with questions, and answers are directly extracted from the text. A
well-known example of an extractive QA dataset is the Stanford Question Answering
Dataset (SQuAD), which includes over 100,000 QA pairs from Wikipedia
articles~\cite{rajpurkar2016squad}. In the case of Icelandic, a language closely related
to Faroese, several QA datasets have been developed.
\citet{snaebjarnarson-einarsson-2022-cross} introduced a cross-lingual open-domain QA
system using machine-translated data, and the Natural Questions in Icelandic, an
extractive QA dataset, which demonstrates approaches applicable to other low-resource
languages such as Faroese~\cite{snaebjarnarson-einarsson-2022-natural}. Similarly,
\citet{skarphedinsson-etal-2023-gameqa} developed a method to gamify QA dataset
creation. However, both approaches relied heavily on human question generation, which
bottlenecked the dataset creation process.

At the time of writing, few benchmark datasets exist for Faroese.
\citet{snaebjarnarson-etal-2023-transfer} introduced named entity
recognition\footnote{\url{https://huggingface.co/datasets/vesteinn/sosialurin-faroese-ner}}
and semantic text similarity
datasets\footnote{\url{https://huggingface.co/datasets/vesteinn/faroese-sts}}. The
FLORES-200 dataset~\cite{nllb2022} is another significant contribution to Faroese
benchmarks, being a multilingual parallel corpus covering over 200 languages, including
Faroese. Additionally, \citet{nielsen2023scandeval} introduced ScaLA-Fo, a linguistic
acceptability dataset for Faroese. Despite these resources, a dedicated Faroese QA
dataset is still lacking, which this work aims to address.

\section{Methodology}
\label{sec:methodology}

\subsection{Generation of Tentative Dataset}
\label{subsec:tentative_dataset}

The process of generating an extractive question-answering dataset begins with several
key components: a vocabulary, a text corpus, a generative model, and specialised
functions for generating questions and answers and for question reformulation. Using
these components, we create a tentative dataset through a two-step process. First, we
apply a QA generation function to our text corpus to create initial QA pairs. Then, we
refine these pairs by rewriting the questions while keeping the answers unchanged.

The QA generation function operates by utilising our generative model to create multiple
questions for each document in the corpus, along with corresponding answers that must be
found verbatim within the source document. To ensure consistency and maintainability, we
implement strict formatting requirements for the model's output. Specifically, we
require the model to generate responses in a structured JSON format, following the
approach described by~\citet{willard2023efficient}. Each output must be a dictionary
containing a ``results'' key, which maps to a list of dictionaries. These inner
dictionaries must contain exactly two keys: ``question'' and ``answer.'' Any outputs
that deviate from this precise format are automatically filtered out of the dataset.

\begin{figure*}[t]
\centering
\begin{tikzpicture}[scale=0.85, transform shape,
    box/.style={draw, rounded corners=5pt, minimum width=3cm, minimum height=1.2cm,
                font=\sffamily\large, text=black!90, line width=0.8pt},
    data/.style={box, fill=green!5, draw=green!60},
    process/.style={box, fill=blue!5, draw=blue!60, minimum width=3.8cm, minimum height=1.4cm},
    note/.style={text width=3.2cm, align=center, font=\sffamily\small, text=black!70},
    arrow/.style={-stealth, thick, draw=gray!50, line width=1pt},
    json/.style={draw=gray!40, rounded corners=5pt, fill=gray!3,
                text width=5.2cm, align=left, font=\ttfamily\scriptsize}
]
% Define spacing
\def\xspacing{4.5}
\def\yspacing{3.2}
% Initial corpus
\node[data] (corpus) at (0,0) {Input Documents};
% QA Generation step
\node[process] (qa_gen) at (\xspacing,0) {\begin{tabular}{c}Generate Questions\\ and Answers\end{tabular}};
% Initial QA pairs
\node[data] (initial_qa) at (2*\xspacing,0) {Initial QA Pairs};
% Question rewriting step
\node[process] (q_rewrite) at (2*\xspacing,-\yspacing) {Rephrase Questions};
% Validation step
\node[process] (validation) at (0,-\yspacing) {Validation};
% Final dataset
\node[data] (final) at (\xspacing,-\yspacing) {Final Dataset};
% Arrows with better spacing
\draw[arrow] (corpus.east) -- (qa_gen.west);
\draw[arrow] (qa_gen.east) -- (initial_qa.west);
\draw[arrow] (initial_qa.south) -- (q_rewrite.north);
\draw[arrow] (q_rewrite.west) -- (final.east);
\draw[arrow] (final.west) -- (validation.east);
% Add explanation with improved positioning
\node[note] (note1) at (\xspacing,1.3)
    {Model creates questions and extracts answers from text};
% Add the process description to the left of the arrow
\path (initial_qa.south) -- (q_rewrite.north) node[note, pos=0.5, anchor=east, xshift=+0.4cm]
    {Model improves question clarity while preserving original answers};
% Enhanced JSON format example
\node[json] at (3*\xspacing+0.9,-1) {
    Example Output Format:\\[0.2em]
    \{\\
    ``results'': [\\
    \hspace{0.3cm}\{\\
    \hspace{0.6cm}``question'': ``What role did Alan\\
    \hspace{0.6cm}Turing play in the development\\
    \hspace{0.6cm}of computer science?'',\\
    \hspace{0.6cm}``answer'': ``Alan Turing laid the\\
    \hspace{0.6cm}theoretical foundation for\\
    \hspace{0.6cm}computer science through his\\
    \hspace{0.6cm}work on computability.''\\
    \hspace{0.3cm}\},\\
    \hspace{0.3cm}\{\\
    \hspace{0.6cm}``question'': ``What was the\\
    \hspace{0.6cm}Turing machine?'',\\
    \hspace{0.6cm}``answer'': ``A mathematical model\\
    \hspace{0.6cm}of computation that manipulates\\
    \hspace{0.6cm}symbols on a tape.''\\
    \hspace{0.3cm}\}\\
    ]\\
    \}
};
\end{tikzpicture}
\caption{Overview of the QA dataset generation pipeline. The system processes input
documents to generate initial QA pairs, followed by a question rewriting phase that
improves clarity while maintaining the original answers. All outputs follow a structured
JSON format to ensure consistency. Note that while the outputs are in Faroese, the
example shown in this figure uses an English example for illustrative purposes.}
\label{fig:qa-generation}
\end{figure*}
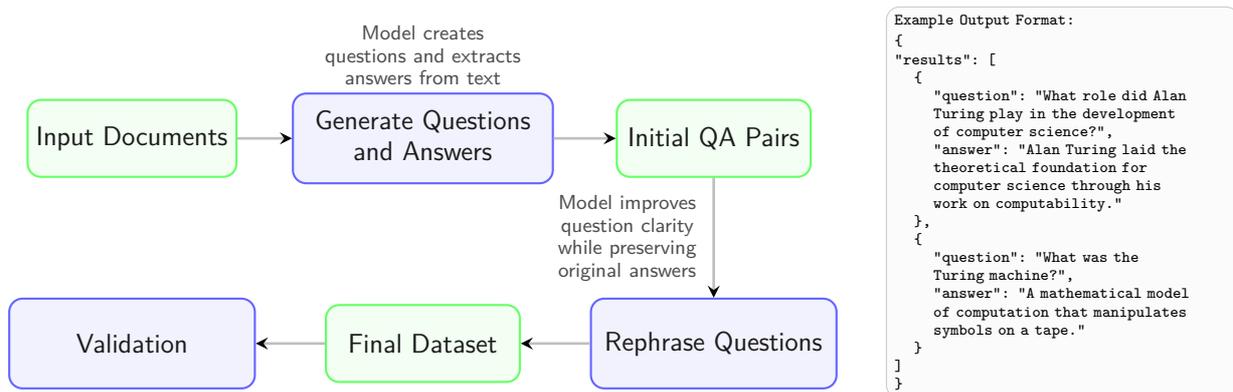

A significant limitation of the initially generated questions lies in their close
adherence to the source documents' original phrasing. These questions often merely
restructure existing statements from the text into interrogative forms, diminishing
their effectiveness as evaluation tools. Consider a document containing the statement
``Jane Smith is an executive and her bike is red.'' The initial generation might produce
``What colour is Jane Smith's bike?''—a question that could be answered through simple
text matching algorithms, requiring minimal linguistic or reasoning capabilities. To
address this limitation, we employ a question reformation process that introduces
additional complexity. By transforming the previous example to ``What colour is the
executive's bicycle?'', we create questions that demand more sophisticated comprehension
abilities, including synonym recognition and multi-hop reasoning in this example. This
reformulation process is implemented through our question-rewriting function, which
utilises the generative model to produce modified questions.

We release our code base implementing this generation process
open-source\footnote{\githuburl}.

\subsection{Manual Filtering of Tentative Dataset}
\label{subsec:manual_filtering}

To ensure high-quality dataset creation, we implemented a human validation phase using a
custom annotation interface built with Gradio~\cite{abid2019gradio}, a Python library
for web-based interfaces. The tool presents annotators with each generated question and
its answer, offering three classification options: \textsc{Correct} (both question and
answer are grammatically and contextually appropriate), \textsc{Incorrect} (question is
grammatically incorrect or contextually inappropriate), and \textsc{Incorrect Answer}
(answer is irrelevant, inaccurate, or grammatically incorrect). An annotator reviews
each QA pair and assigns the appropriate classification, ensuring linguistic quality and
filtering out inadequate samples. The annotation tool is available
open-source\footnote{\appurl}.

\subsection{Annotation Guidelines}

This section outlines the complete annotation guidelines for evaluating QA pairs in
Faroese. The annotator will follow a three-tier classification system when analysing
each sample.

\paragraph{Tier 1: Grammatical Assessment}
The annotator should begin by evaluating the grammatical correctness of both the
question and answer in Faroese. The annotator must check for proper agreement between
subjects and verbs, correct case marking on nouns and pronouns, standard Faroese word
order, accurate spelling and so on. If any grammatical errors are found in the question,
the annotator should mark the entire sample as \textsc{Incorrect}. If the grammatical
errors only appear in the answer, the sample should be marked as \textsc{Incorrect
Answer}.

\paragraph{Tier 2: Semantic and Contextual Assessment}
After confirming grammatical correctness, the annotator should examine the relationship
between the question and answer, as well as their connection to the source text. The
answer must directly address the question being asked. Additionally, the annotator
should ensure the answer demonstrates logical consistency within its context. If any
issues with relevance, accuracy, or consistency are found, the sample should be marked
as \textsc{Incorrect Answer}.

\paragraph{Tier 3: Final Classification}
When a sample passes both the grammatical and semantic assessments, the annotator should
mark it as \textsc{Correct}. The annotator will also be asked to correct a selection of
questions marked as \textsc{Incorrect}. When performing these corrections, the annotator
should focus only on samples where the question itself contains errors, not the answer.
This is crucial because modifying answers would compromise the extractive nature of the
QA task, as answers should appear verbatim in the source text.

\paragraph{Quality Control Process}
All samples marked as \textsc{Correct} can undergo a secondary review by another
annotator who is also a native Faroese speaker. This second annotator will apply the
same three-tier evaluation process described above.

\section{Faroese Setup}
\label{sec:faroese_setup}

We applied our methodology (Section~\ref{sec:methodology}) to the Faroese
Wikipedia\footnote{This dump: \wikiurl.} as the text source and
\texttt{gpt-4-turbo-2024-04-09} \cite{openai2023newmodels} as the generative model,
selected for its top performance on Faroese tasks in the ScandEval
benchmark~\cite{nielsen2023scandeval,nielsen2024encoder}. To ensure non-trivial
contexts, only articles with over 1,000 characters were included, i.e., 1675 articles in
total and 655 articles used for the validated dataset. We set the model temperature to
1.0 and generated a maximum of 1,024 tokens, with a consistent random seed (4242) to
maintain reproducibility. The system prompt we use is the following:

\begin{quote}
\small
You are a helpful Faroese question answering dataset generator. The only language you know is Faroese.
\end{quote}

While we did not explore Faroese-language prompting or prompt variations in this study,
such modifications could potentially improve the effectiveness of our approach. As our
primary focus was developing a question-answering dataset for Faroese, we leave prompt
optimisation for future work. The following prompt was used for generating QA pairs:

\begin{quote}
\small
The following is a Wikipedia article in Faroese.

<article>\\
\{article\}\\
</article>

Generate 2 to 10 questions about the article, depending on the length of the article, all of which answered in the article.

You also have to supply answers to the questions, and the answers have to appear exactly as written in the article (including same casing).

The answers should only contain the answers themselves, and not the surrounding sentence - keep the answers as short as possible.

The answers have to be different from each other.

All your questions and answers must be in Faroese.

Your answer must be a JSON dictionary with the key ``results'', with the value being a list of dictionaries having keys ``question'' and ``answer''.
\end{quote}

Lastly, we use the following prompt to re-write the questions:

\begin{quote}
\small
The following is a Faroese question.

<question>\\
\{question\}\\
</question>

Re-write the question, preserving the meaning, using synonyms or a different (valid) word order.

Your question must be in Faroese.

Your answer must be a JSON dictionary with the key ``question''.
\end{quote}

In both prompts, we replace \{article\} and \{question\} with the actual Wikipedia
article and the generated question, respectively.

\section{The Dataset}
\label{sec:thedataset}

\subsection{Format}
\label{subsec:format}

The validated QA pairs are stored in a structured format, with each entry containing a
unique identifier (\texttt{id}), the source article's URL (\texttt{url}), the article
title (\texttt{title}), the full text (\texttt{context}), the generated and rephrased
question (\texttt{question}), and an answers dictionary (\texttt{answers}) that includes
the answer text and its character index (\texttt{answer\_start}) within the context.
This structure ensures compatibility with standard extractive QA formats like SQuAD
\cite{rajpurkar2016squad}, enabling seamless integration with existing NLP frameworks
and models.

\subsection{Statistics}
\label{subsec:statistics}

The tentative dataset in our Faroese case consisted of \textbf{10,001} samples, which
were randomly selected from the Wikipedia articles meeting our length criteria (>1,000
characters). From these samples, \textbf{4,130} were annotated by a human annotator. Out
of the annotated samples, \textbf{1,759} were annotated as \textsc{Correct},
\textbf{1,908} were \textsc{Incorrect}\footnote{While more than half of all generated
question/answer pairs were marked as incorrect, we release the full dataset to enable
researchers to study GPT-4's error patterns in Faroese.} and \textbf{222} had an
\textsc{Incorrect Answer}. While the initial validation was performed by a single
annotator, we conducted a second validation phase specifically for the samples marked as
\textsc{Correct}, where these samples were evenly split between two annotators: the
original annotator and a second native Faroese speaker. During this step, \textbf{41}
out of the \textbf{1,759} \textsc{Correct} samples were found to have been incorrectly
labelled as \textsc{Correct} by the annotator, which was then corrected. Additionally,
\textbf{241} samples have the label \textsc{Corrected} where the original question has
been corrected by the human annotator (this includes the 41 incorrectly labelled samples
which were corrected). These corrected samples are intended to both measure and mitigate
potential biases introduced by GPT-4-turbo during the initial sample creation. By
comparing model performance on the corrected versus uncorrected samples of the dataset,
we can assess whether the model exhibits any bias toward its own generated questions.

\subsection{Dataset Versions}
\label{subsec:versions}
We are releasing three versions of the FoQA dataset on the Hugging Face
Hub\footnote{Available at \hfurl.}. The format of the dataset is compatible with
standard extractive QA formats like SQuAD. The primary version, \textbf{default},
contains 2,000 validated examples (comprising 1,759 initially correct examples and 241
examples that were corrected during review), including 848 for training, 128 for
validation, and 1,024 for testing, with shortened contexts for improved usability. The
second version, \textbf{all-samples}, includes all 10,001 examples from the initial
dataset, retaining full, unshortened contexts, even those that were rejected or not
validated. The final version, \textbf{incorrect-samples}, comprises 2,395 examples that
were rejected during the manual review process.

\subsubsection{Question Types}
We used the \texttt{gpt-4o-2024-05-13} model from OpenAI to annotate the questions into
categories and we used the following system prompt:

\begin{quote}
    \small
    Categorize the question (written in Faroese) based on the type of question it is.
    The question types are ``time'' for questions that ask about the time of something,
    ``place'' if they ask for a place, ``people'' if they ask about a person, ``object''
    if they ask about an object or a non-person entity. If the question does not fit any
    of these categories, respond with ``other''.
\end{quote}

Most questions received the people label (679, 33.95\%), followed by object (516,
25.80\%), time (367, 18.35\%), place (290, 14.50\%) and other (148, 7.40\%).

To assess the quality of the automatic question categorisation, the annotator manually
validated 200 randomly sampled questions from the dataset. The validation methodology
included assigning binary scores: 1 for correct categorization and 0 for incorrect
categorisation. The validation followed an inclusive approach, accepting multiple valid
category assignments where applicable. For instance, questions about a person's
birthplace (e.g., ``Where was Turið Sigurðardóttir born?'') were considered correctly
categorised if labelled as either ``person'' or ``place,'' as both categories are
contextually relevant to the question's intent. This flexible validation framework
acknowledges the inherent ambiguity in question categorisation, where multiple
interpretations may be equally valid.

The manual validation revealed an error rate of 7.5\% (15 incorrect categorisations out
of 200 validated samples), suggesting that the GPT-4o categorisation system achieved
92.5\% accuracy on the validated subset.

The annotator also conducted a qualitative error type analysis. Here it was found that
common error types in the QA dataset include grammatical gender mistakes, such as using
neuter instead of masculine forms in questions about pool length (e.g., ``Hvussu langur
er svimjihylið.NEUT í kappingunum''). Incorrect phrasing surrounding years, like
omitting the preposition ``í'' (in) when asking about dates (e.g., ``Hvørjum ári doyði
Stephen Hawking?''), is also prevalent. Icelandicisms appear as words that are partially
or fully Icelandic (e.g., the use of ``hraði'' (speed) inflected as a Faroese noun in
``Hvør er hraðin á jørðini í kilometrum hvønn tíma?''). The questions and answers also
contained errors in punctuation, spelling, and capitalization, as seen in the improper
capitalization of ``Smyril'' (merlin) when referring to the bird rather than the ferry
(e.g., ``Hvat ger Smyril?''). Lastly, some incorrect terms are used consistently (e.g.,
``høvuðsbýur'' (main city) used instead of ``høvuðsstaður'' (capital) when asking about
capital cities).

\section{Evaluation}
\label{sec:evaluation}

We evaluated several models on the dataset. Since we ensured that all answers appear
exactly as in the documents, this allows us to evaluate both encoder models and decoder
models on the dataset. We evaluate both Faroese and massively multilingual models on
FoQA, the results of which can be found in Table~\ref{tab:results}.

\begin{savenotes}
    \begin{table}
        \centering
        \small
        \begin{tabular}{l|cc}
            \textbf{Model Name} & \textbf{F1 Score} & \textbf{Exact Match} \\
            \midrule
            GPT-4-turbo\footnote{Full OpenAI model ID: \texttt{gpt-4-1106-preview}} & 77.6 ± 1.0 & 55.6 ± 1.8 \\
            GPT-4o\footnote{Full OpenAI model ID: \texttt{gpt-4o-2024-05-13}} & 77.1 ± 1.0 & 54.1 ± 1.6 \\
            GPT-4o-mini\footnote{Full OpenAI model ID: \texttt{gpt-4o-mini-2024-07-18}} & 75.2 ± 1.0 & 51.2 ± 1.5 \\
            Llama-3.1-8B\footnote{\url{https://hf.co/meta-llama/Llama-3.1-8B}} & 73.6 ± 1.2 & 51.9 ± 1.5 \\
            GPT-SW3-6.7B\footnote{\url{https://hf.co/AI-Sweden-Models/gpt-sw3-6.7b-v2}} & 63.4 ± 2.2 & 45.2 ± 2.1 \\
            Mistral-7B\footnote{\url{https://hf.co/mistralai/Mistral-7B-v0.3}} & 62.4 ± 1.7 & 45.0 ± 1.6 \\
            FoBERT\footnote{\url{https://hf.co/vesteinn/FoBERT}} & 36.0 ± 1.7 & 26.8 ± 1.5 \\
            mDeBERTa-v3\footnote{\url{https://hf.co/microsoft/mdeberta-v3-base}} & 30.6 ± 1.6 & 21.0 ± 1.2 \\
            ScandiBERT\footnote{\url{https://hf.co/vesteinn/ScandiBERT-no-faroese}} & 30.9 ± 2.7 & 21.9 ± 2.3 \\
        \end{tabular}
        \caption{Evaluation results on FoQA according to F1 scores and exact match.}
        \label{tab:results}
    \end{table}
\end{savenotes}

We also evaluated the model used to generate the dataset,
\texttt{gpt-4-turbo-2024-04-09}, on the corrected samples, before and after the
correction. This was to test whether the model is biased towards its own generated
questions, or whether it generalises to the corrected ones as well. Surprisingly, the
model ended up performing significantly better\footnote{$p=0.0007$ for F1-score and
$p=0.0185$ for exact match, using a two-tailed t-test.} on the corrected samples, rather
than the samples it had generated itself.

\section{Discussion and Future Work}

Our evaluation of the FoQA dataset reveals insights into the performance of various
language models on Faroese QA tasks. GPT-4-turbo and GPT-4o achieved the highest
performance scores in our evaluation, though further research would be needed to
understand whether this indicates genuine Faroese language comprehension or other
factors like strong general question-answering capabilities. This finding suggests
promising directions for low-resource language processing, while highlighting the need
for more detailed investigation into how these models handle Faroese specifically.

An important observation from our annotator indicates that most errors in the generated
questions were grammatical in nature rather than contextual. This suggests a need for
dedicated benchmarks specifically measuring grammatical correctness of LLMs in Faroese,
which would complement FoQA's focus on QA capabilities.

Early question answering datasets like SQuAD faced criticism that their questions were
too simplistic, often directly mirroring the source text structure. Later datasets like
TyDi QA~\cite{10.1162/tacl_a_00317} and Natural Questions in
Icelandic~\cite{snaebjarnarson-einarsson-2022-natural} addressed this by having
annotators create natural questions first, which were later matched to source material.
This approach prevented the tight coupling between question phrasing and source text
that can make questions artificially easy. Following this insight, we implemented
question rephrasing in our methodology. However, we acknowledge that we did not
specifically measure performance differences between original and rephrased questions,
which would require separate evaluation sets.

We found that encoder models like mDeBERTa-v3~\cite{he2021debertav3, he2021deberta},
FoBERT and ScandiBERT~\cite{snaebjarnarson-einarsson-2022-cross} perform significantly
worse than the decoder models, but that could simply be explained by the fact that these
models differ in sizes by several orders of magnitude. A controlled experiment will need
to reveal whether architectural choices are the real cause for difference in performance
or whether it is due to other reasons such as parameter count. A performance gap has
been observed between encoder-type models and decoder-type models across other languages
and \cite{nielsen2024encoder} suggests that certain architectures may be inherently
better suited for specific language processing tasks.

For future work, we propose evaluating larger open models, such as 70B parameter models
and even larger ones like Llama 3.1 405B~\cite{dubey2024llama3herdmodels}. Additionally,
assessing the performance of Claude 3.5 Sonnet~\cite{claude_sonnet_3_5} would be
valuable, given its strong performance on Icelandic NLP
tasks\footnote{\url{https://huggingface.co/spaces/mideind/icelandic-llm-leaderboard}}
since Icelandic is a language closely related to Faroese.

\section{Conclusion}
\label{sec:conclusion}

We introduced FoQA, the first Faroese extractive question-answering dataset, containing
2,000 QA pairs. All samples underwent initial validation by one annotator, followed by a
second validation phase where the correct samples were split equally between the
original annotator and a second annotator. Our evaluation reveals significant
performance gaps between decoder-based LLMs and encoder models, with GPT-4-turbo
achieving the highest F1 score of 77.6, while encoder models like mDeBERTa-v3 and
ScandiBERT scored around 30. Notably, our analysis of question types shows a diverse
distribution across categories, with people-related questions comprising the largest
portion at 33.95\%. The dataset's manual validation process identified common error
patterns, including grammatical gender mistakes and Icelandicisms, providing valuable
insights for future Faroese language model development. The FoQA dataset serves as
valuable benchmark for evaluating Faroese language understanding. Additionally, our
contributions include a semi-automated methodology for creating extractive QA datasets
for low-resource languages.

\section*{Limitations}

A significant limitation of our dataset is that our current annotation process does not
differentiate between grammatical errors and contextual errors in the generated
questions. This granular error categorisation would provide valuable insights for
improving model performance and understanding specific challenges in Faroese language
generation.

The use of GPT-4-turbo for dataset generation introduces potential biases in the
linguistic patterns of the generated text. Despite native speaker validation, there
remains a risk that the generated questions may not fully capture natural Faroese
language patterns and could subtly reflect machine-generated language characteristics.

Our methodology relied on a single annotator for the initial validation phase, which
means we could not perform traditional inter-annotator agreement measurements. While
this limitation was intentional, as our approach aimed to demonstrate the feasibility of
creating useful datasets with minimal human resources, it does impact our ability to
measure annotation consistency quantitatively. Another limitation is our lack of
evaluation on the non-rephrased questions. This missing comparison makes it difficult to
quantify the impact of our question rephrasing strategy and determine whether it
actually increased question difficulty as intended.

Furthermore, Faroese Wikipedia, while a valuable resource, is relatively small and
occasionally contains ungrammatical content due to a limited pool of contributors. This
occasionally led to incorrect-answer errors, since the answers are extracted directly
from the source text. And lastly, the current size of 2,000 validated QA pairs, while a
solid starting point, is relatively small compared to QA datasets for high-resource
languages, which may limit its capacity to train or fine-tune LLMs effectively.

\section*{Ethical Statement}

The creation of language resources for low-resource languages like Faroese raises
important ethical considerations, particularly when utilising LLMs. Our dataset
generation process involved processing approximately 1,675 Faroese Wikipedia articles
through GPT-4-turbo. While this automated approach enabled efficient initial data
generation, we acknowledge the computational resources required and their environmental
impact, and we conservatively estimate that the processing spanned 48 GPU hours. We note
that OpenAI’s infrastructure runs on Azure, and Azure will be running on 100\% renewable
energy by 2025 and has been carbon neutral since 2012\footnote{See more information on
\href{https://azure.microsoft.com/en-gb/explore/global-infrastructure/sustainability}{Azure's
sustainability page}.}.

A primary ethical concern in using LLMs for low-resource language content generation is
the potential introduction of non-native language patterns and cultural
misrepresentations. This risk is particularly relevant for Faroese, where preserving
authentic linguistic patterns and cultural context is crucial. To address these
concerns, we implemented a comprehensive validation protocol requiring native speaker
review of all generated content. This human-in-the-loop approach helped identify and
correct systematic errors while ensuring linguistic authenticity.

To maximise the dataset's benefit to the Faroese language technology community, we have
made it freely available under an open-source license. We are committed to ongoing
maintenance and error correction, ensuring the dataset remains a valuable resource for
Faroese language technology development while maintaining high standards of linguistic
quality and cultural authenticity.

\section*{Acknowledgments}
\label{sec:acknowledgments}

AS was supported by the European Commission under grant agreement no. 101135671. We
thank the reviewers for their constructive and helpful comments, which have
significantly improved the quality and clarity of our manuscript.

\newpage
\bibliographystyle{acl_natbib}
\bibliography{foqa}

\end{document}